\documentclass[letterpaper]{article} %DO NOT CHANGE THIS
\usepackage{aaai19}  %Required
\usepackage{booktabs} % For formal tables
\usepackage{bm}
\usepackage{algorithm,algorithmic}
\usepackage{amsfonts}
\usepackage{mathtools}
\usepackage{subcaption}
\usepackage{times}  %Required
\usepackage{helvet}  %Required
\usepackage{courier}  %Required
\usepackage{url}  %Required
\usepackage{graphicx}  %Required
\begin{document}
 	\title{Enhancing Evolutionary Conversion Rate Optimization\\ via Multi-armed Bandit Algorithms}
 	\author{Xin Qiu \textsuperscript{1}, Risto Miikkulainen \textsuperscript{1, 2}\\
 		\textsuperscript{1} Sentient Technologies, Inc\\
 		\textsuperscript{2} The University of Texas at Austin\\
 	}

\maketitle

\begin{abstract}
Conversion rate optimization means designing web interfaces such that more visitors perform a desired action (such as register or purchase) on the site. One promising approach, implemented in Sentient Ascend, is to optimize the design using evolutionary algorithms, evaluating each candidate design online with actual visitors. Because such evaluations are costly and noisy, several challenges emerge: How can available visitor traffic be used most efficiently? How can good solutions be identified most reliably? How can a high conversion rate be maintained during optimization? This paper proposes a new technique to address these issues. Traffic is allocated to candidate solutions using a multi-armed bandit algorithm, using more traffic on those evaluations that are most useful. In a best-arm identification mode, the best candidate can be identified reliably at the end of evolution, and in a campaign mode, the overall conversion rate can be optimized throughout the entire evolution process. Multi-armed bandit algorithms thus improve performance and reliability of machine discovery in noisy real-world environments. 
\end{abstract}

\section{Introduction}
Conversion rate optimization (CRO) is an emerging field of applied AI \cite{Khalid11}. Web interface designs are optimized to increase the percentage of visitors who perform a desired action such as making a purchase, registering a new account, or clicking on a desired link. The true conversion rate of a web interface is unknown, but can be estimated via a number of user interactions, which are usually noisy. CRO is therefore a challenging application of optimization under uncertainty.

Recently, a new technology for CRO was developed in a system called Sentient Ascend \cite{Miikkulainen2017,miikkulainen:iaai18}.  Each website design is represented as a genome, and evolutionary algorithms (EAs) are used to search for designs that convert well. Evolutionary CRO provides considerable advantages over traditional A/B or multivariant testing: Exploration in EA covers a large design space; evolution discovers and utilizes effective interactions among variables; optimization of website design is fully automated.

Although this approach leads to impressive improvements over human design \cite{Miikkulainen2017,miikkulainen:iaai18}, several open issues remain in evolutionary CRO. First, candidate designs are expensive to evaluate, and traffic is often wasted on bad designs. Second, only weak statistical evidence is available to select a winner design, reducing reliability of the optimization outcome. Third, in some cases, the target is to maintain a high overall conversion rate during the optimization process instead of identifying a single best design at the end of the optimization.

To overcome these issues, this paper proposes augmenting EAs with multi-armed bandit (MAB) algorithms \cite{AgrawalG12,Audibert2010,Auer2002}. First, a new framework for traffic allocation during fitness evaluation is developed based on MAB, called MAB-EA. This framework aims at reducing the evaluation cost while maintaining the optimization performance. Second, an enhanced variant of MAB-EA is designed to select the winner reliably. The main idea is to include an addtional verification phase, based on MAB algorithms, to the end of the evolution process. Third, another variant of MAB-EA is developed by introducing a new concept called asynchronous statistics into MAB algorithms. The new variant is particularly well suited for situations where overall conversion rate during optimization needs to be maximized. Empirical studies with simulated traffic in Sentient Ascend demonstrate that the MAB techniques are effective in Evolutionary CRO. The techniques proposed in the paper are general, however, and it should be possible to adapt them to other optimization problems in uncertain environments.

The remainder of this paper is organized as follows. First, background knowledge regarding evolutionary CRO and MAB algorithms is provided. The technical details of the new approaches are then explained, and the underlying rationale is discussed. After that, the proposed techniques are evaluated experimentally with Sentient Ascend on simulated traffic. Evaluation of the results and suggestions for future work conclude the paper.

\section{Background}
This section describes the basic concepts and existing challenges in evolutionary CRO. A brief introduction of MAB problem and representative MAB algorithms is then provided.
\subsection{Evolutionary Conversion Rate Optimization}
EA is a population-based metaheuristic inspired by natural evolution process. Each individual (genome) in the population represents a single solution to the optimization problem, and these individuals will evolve through crossover, mutation and survival selection iteratively. The most significant advantage of EAs is that they do not make any assumption about the underlying landscape of the optimization problems, thereby leading to exceptional ability in finding good solutions to mathematically intractable problems \cite{Nature2015}. 

In evolutionary CRO \cite{Miikkulainen2017,miikkulainen:iaai18}, each genome represents a web interface design. The search space is pre-defined by the web designer. For each such space, the designer specifies the elements of the interface and values that they can take. For instance in a landing page, logo size, header image, button color, content order are such elements, and they can each take on 2-4 values. Evolutionary CRO searches for good designs in the space of possible combinations of these values, which often number in millions. In each generation, all the genomes will be evaluated with a fixed number of user interactions, and the conversion rates during evalution will be used as the fitnesses for each genome. Fitness-proportionate selection is then used to select parent genomes, and traditional genetic operations such as crossover (recombination of the elements in two parent genomes) and mutation (randomly changing one element in the offspring genome) are performed to generate offspring candidates. The same process will be repeated generation by generation until the termination criterion is met, which usually means reaching a fixed number of user interactions. In a typical winner-selection application, the winning design is then selected among the best candidates, with an estimate of its future performance. In campaign-mode application, there is no winner but performance is measured by the overall convergence rate throughout the entire experiment.
\subsubsection{Challenges in Real-World Evolutionary CRO}
When the Evolutionary CRO methods were taken out of the laboratory and into the real world application, it became clear that there were new and interesting challenges that needed to be met. First, in the original Evolutionary CRO framework \cite{Miikkulainen2017,miikkulainen:iaai18}, the evaluation of each candidate is performed in a static fashion: A fixed amount of traffic is allocated to each web design. This means even if a candidate is clearly bad based on a few visits, the system currently gives it the same amount of traffic as for good ones. A large amount of real traffic may be wasted by bad candidates, leading to more expensive evaluations. Second, during the normal evolutionary process, only weak statistical evidence is obtained. Therefore, there is a multiple hypotheses problem, i.e. the winner candidate is most likely not the one with the best true conversion rate, but one that got lucky with evaluations. Third, the current evolutionary CRO technique is designed to identify a good candidate at the end of optimization. However, in some scenaria, the goal for CRO is to make the overall conversion rate during optimization as high as possible. With uniform traffic allocation, bad candidates are tested as much as good ones, thereby reducing the overall conversion rate. To address these issues, this paper presents a new technique that takes advantage of MAB algorithms in evolutionary CRO.

\subsection{Multi-armed Bandit Algorithms}
This subsection explains the definition of multi-armed bandit problem, and introduces three representative multi-armed bandit algorithms used in this paper.
\subsubsection{Multi-armed Bandit Problem}
In MAB problem, a slot machine with multiple arms is given, and the gambler has to decide which arms to pull, how many times to pull each arm, and in which order to pull them \cite{Richard92}. The most common is the stochastic MAB problem, which is parameterized by the number of arms $K$, the number of rounds $n$, and $K$ fixed but unknown reward distributions $\nu_{1}, \nu_{2}, \ldots, \nu_{K}$ associated with arm 1, arm 2, $\ldots,$ arm $K$,  respectively. For $t = 1, 2, \ldots, n$, at round $t$, the agent (gambler) chooses an arm $I_{t}$ from the set of arms $\{1, 2, \dots, K\}$ to pull, and observes a reward sampled from $\nu_{I_{t}}$. Each reward sample is independent from the past actions and observations. The CRO problem is a special case (called Bernoulli bandit) of the general stochastic MAB problem: the reward for each pull is either 1 or 0 (converted or not in CRO), and for arm $i$ the probability of success (reward = 1) is $p_i$, which equals to its true conversion rate.

An algorithm for the stochastic MAB problem must decide which arm to pull at each round $t$, based on the outcomes of the previous $t - 1$ pulls. In the classical MAB problem, the goal is to maximize the cumulative sum of rewards over the $n$ rounds \cite{robbins1952,Auer2002}. Since the agent has no prior knowledge about the reward distributions, it needs to explore the different arms, and at the same time, exploit the seemingly most rewarding arms \cite{Audibert2010}. This goal aligns with the campaign-mode application in CRO. For clarity of statement, we call this type of problem the classical stochastic MAB problem. Another target for stochastic MAB problem is to output a recommended arm after a given number of pulls. The performance of the MAB algorithm is only evaluated by the average payoff of that recommended arm. This goal aligns with the winner-selection application in CRO. This is called pure exploration problem \cite{Bubeck2009}.
\subsubsection{UCB Algorithm}
Upper Confidence Bound (UCB) algorithm is arguably the most popular approach for solving classical MAB problems due to its good theoretical guarantees \cite{AgrawalG12}. The principle behind UCB is optimism in the face of uncertainty \cite{KAMIURA201725}. Generally, UCB constructs an optimistic guess on the potential reward of each arm, and pulls the arm with the highest guess. Among the UCB family of algorithms, UCB1 \cite{Auer2002} is a simple yet efficient variant that can be directly applied to Bernoulli Bandits. The optimistic guess in UCB1 is in the form of an upper confidence bound derived from the Chernoff-Hoeffding inequality \cite{Chernoff1952,Hoeffding1963}. Algorithm 1 shows the basic steps of UCB1 algorithm.
\begin{algorithm}[tb]
	\caption{UCB1 Algorithm}
	\label{pseudo_UCB1}
	\begin{algorithmic}[1]
		\REQUIRE ${}$
		\\ $K$: Total number of arms
		
		\FOR{$i = 1$ to $K$}
		\STATE Pull arm $i$, and observe reward $X_{i,0}$
		\STATE $\hat{x}_i = X_{i,0}$
		\STATE $n_i = 1$
		\ENDFOR
		\FOR{$t = K+1, K+2, \ldots$}
		\STATE Pull arm $i_{\mathrm{max}} := \mathrm{arg max}_i\hat{x}_i + \sqrt{2 \log{(t)} / n_i}$, and observe reward $X_{i_{\mathrm{max}}, t}$
		\STATE $\hat{x}_{i_{\mathrm{max}}} = \frac{\hat{x}_{i_{\mathrm{max}}} \times n_{i_{\mathrm{max}}} + X_{i_{\mathrm{max}}, t}}{n_{i_{\mathrm{max}}} + 1}$
		\STATE $n_{i_{\mathrm{max}}} = n_{i_{\mathrm{max}}} + 1$
		\ENDFOR
	\end{algorithmic}
\end{algorithm} 
\subsubsection{Thompson Sampling}
Except for UCB, Thompson Sampling (TS) \cite{Thompson1933} is another good alternative MAB algorithm for the classical stochastic MAB problem. The idea is to assume a simple prior distribution on the parameters of the reward distribution of every arm, and at each round, play an arm according to its posterior probability of being the best arm \cite{AgrawalG12}. The effectiveness of TS has been empirically demonstrated by several studies \cite{Ole2010,Scott2010,Chapelle2011}, and the asymptotic optimality of TS has been theoretically proved for Bernoulli bandits \cite{Kaufmann2012,AgrawalG12}. TS for Bernoulli bandits utilizes beta distribution as priors, i.e., a family of continuous probability distributions on the interval $[0, 1]$ parameterized by two positive shape parameters, denoted by $\alpha$ and $\beta$. The mean of $\mathrm{Beta}(\alpha, \beta)$ is $\frac{\alpha}{\alpha + \beta}$, and higher $\alpha, \beta$ lead to tighter concentration of $\mathrm{Beta}(\alpha, \beta)$ around the mean. TS initially assumes each arm $i$ to have prior reward distribution $\mathrm{Beta}(1, 1)$, which is equivalent to uniform distribution on $[0, 1]$. At round $t$, after having observed $S_i$ successes (reward = 1) and $F_i$ failures (reward = 0) in $S_i + F_i$ pulls for arm $i$, the reward distribution of arm $i$ will be updated as $\mathrm{Beta}(S_i + 1, F_i + 1)$. The algorithm then samples from these updated reward distributions, and selects the next arm to pull according to the sampled reward. Algorithm 2 describes the detailed procedure of TS.
\begin{algorithm}[tb]
	\caption{Thompson Sampling for Bernoulli Bandits}
	\label{pseudo_TS}
	\begin{algorithmic}[1]
		\REQUIRE ${}$
		\\ $K$: Total number of arms
		
		\FOR{$i = 1$ to $K$}
		\STATE $S_i = 0$, $F_i = 0$
		\ENDFOR
		\FOR{$t = 1, 2, \ldots$}
		\FOR{$i = 1$ to $K$}
		\STATE Sample $\theta_{i, t}$ from $\mathrm{Beta}(S_i + 1, F_i + 1)$
		\ENDFOR
		\STATE Pull arm $i_{\mathrm{max}} := \mathrm{arg max}_i\theta_{i, t}$, and observe reward $X_{i_{\mathrm{max}}, t}$
		\IF{$X_{i_{\mathrm{max}}, t} = 1$}
		\STATE  $S_i = S_i + 1$
		\ELSE
		\STATE $F_i = F_i + 1$
		\ENDIF
		\ENDFOR
	\end{algorithmic}
\end{algorithm}
\subsubsection{Successive Rejects Algorithm}
Among many existing algorithms for solving the pure exploration problem, Successive Rejects (SR) algorithm stands out in that it is parameter-free and independent of the scaling of the rewards \cite{Audibert2010}. The main task for SR algorithm is to identify the best arm (the arm with truly best mean reward) after a fixed number of pulls. Suppose we are given $K$ arms and $n$ pulls. First the SR algorithm divides the $n$ pulls into $K - 1$ phases. At the end of each phase, the arm with the lowest empirical mean reward will be discarded. During each phase, each arm that has not been discarded yet will be pulled for equal number of times. The only surviving arm after $K - 1$ phases, $J_n$, will be recommended as the best arm. The SR algorithm is essentially optimal because the regret (difference between the mean rewards of identified best arm and true best arm) decreases exponentially at a rate which is, up to a logarithmic factor, the best possible \cite{Audibert2010}. The details of SR are described in Algorithm 3.
\begin{algorithm}[tb]
	\caption{Successive Rejects Algorithm}
	\label{pseudo_SR}
	\begin{algorithmic}[1]
		\REQUIRE ${}$
		\\ $K$: Total number of arms
		\\ $n$: Total number of pulls
		
		\ENSURE ${}$
		\\ Best arm $J_n$
		\STATE $A_1 = \{1, \ldots, K\}$, $\overline{\mathrm{log}}(K) = \frac{1}{2} + \sum_{i = 2}^{K}\frac{1}{i}$, $n_0 = 0$
		\FOR{$k = 1$ to $K - 1$}
		\STATE $n_{k} = \lceil \frac{1}{\overline{\mathrm{log}}(K)}\frac{n - K}{K + 1 - k}\rceil$
		\FOR{$i \in A_k$}
		\STATE Pull arm $i$ for $n_k - n_{k - 1}$ rounds
		\ENDFOR
		\STATE $A_{k + 1} = A_k \backslash \mathrm{arg min}_{i \in A_k}\hat{X}_{i, n_k}$, where $\hat{X}_{i, n_k}$ is the average reward for arm $i$ after $n_k$ pulls
		\ENDFOR
		\STATE Let $J_n$ be the unique element of $A_K$
	\end{algorithmic}
\end{algorithm}
\section{Methodology}
This section describes the algorithmic details of the proposed approaches and mechanisms. A basic framework combining Evolutionary CRO technique with MAB algorithm is presented first, then two enhanced variants are developed for tackling different use cases, namely, Best Arm Identification and Campaign mode.
\subsection{MAB-EA}
The first goal of this work is to develop a new framework that allocates traffic dynamically in a more efficient way. MAB algorithms are well suited for this role. Each candidate web design can be regarded as an arm, and each visit to the website is equal to a pull. The reward of each visit to a single web design is assumed to follow an unknown but fixed Bernoulli distribution. The probability of getting reward 1 (the visited user is successfully converted) is $p$ and the probability of getting reward 0 (the visited user is not converted) is $1-p$, where $p$ is the true conversion rate of that web design. Given a fixed budget of traffic (number of visits) for each generation, a Bernoulli MAB algorithm will then be invoked to allocate traffic to the current candidates. The fitness of each candidate is equivalent to its number of successful conversions divided by its total visits (both numbers are counted within the current generation). Based on these fitnesses, standard EA operations such as parent selection, crossover, mutation and survival selection will be conducted to generate the population for next generation. Algorithm 4 depicts the procedure of the proposed framework, namely MAB-EA.
\begin{algorithm}[tb]
	\caption{MAB-EA}
	\label{pseudo_BasicFramework}
	\begin{algorithmic}[1]
		\REQUIRE ${}$
		\\ $K$: Population size, $G_{\mathrm{max}}$: Total number of generations
		\\ $T$: Number of website visits for each generation
		\\ $C_e$: Percentage for elites, $C_p$: Percentage for parents
		\\ $P_1$: Initial population, $C_m$: Mutation probability
		\STATE $D = P_1$, $D$ is the archive for storing evaluated candidates
		\FOR{$g = 1$ to $G_{\mathrm{max}}$}
		\STATE Perform MAB algorithm on $P_g$ with a traffic budget of $T$, record the number of conversions $s_i$ and number of visits $n_i$ within current generation for each candidate
		\FOR{$i = 1$ to $K$}
		\STATE Set fitness for candidate $i$ as $f_i = s_i/n_i$
		\ENDFOR
		\STATE Create elite pool $E_g$ as the best $C_e$ percentile candidates in current generation
		\STATE Create parent pool $A_g$ as the best $C_p$ percentile candidates in current generation
		\STATE Initialize offspring pool $O_g$ as empty
		\WHILE{Size of $E_g + O_g$ is less than $K$}
		\STATE Perform fitness-proportionate selection on $A_g$ to pick 2 parent candidates
		\STATE Perform uniform crossover between the two parents to generate an offspring
		\STATE Perform mutation operation on the offspring, each element of the offspring will have $C_m$ probability to be randomly altered
		\IF{the offspring is not in $D$}
		\STATE Add the offspring to $O_g$
		\STATE Add the offspring to $D$
		\ENDIF
		\ENDWHILE
		\STATE $P_{g+1} = E_g + O_g$
		\ENDFOR
	\end{algorithmic}
\end{algorithm}

Note that the goals of the MAB algorithm and the evaluation phase in EA are inherently different: MAB algorithm only cares about identifying the good arms efficiently, whereas evaluation phase in EA aims at estimating the fitnesses of all arms. In spite of this fact, MAB algorithm should not impair the optimization performance significantly, and may even improve it. As shown in Algorithm 4, elite candidates play an important role in both parent selection and survival selection. Since MAB algorithms allocate more traffic to those promising candidates, their fitnesses are actually more reliably estimated than those of the least promising candidates. Selection mechanisms relying on good candidates are therefore further enhanced. The proposed framework is expected to significantly increase the overall conversion rate during evolution without sacrificing the overall optimization performance.
\subsection{Best Arm Identification Mode}
One classical task for CRO is to identify a single best design that can be delivered to website owner for long-term use. The reliability of the optimization outcome therefore becomes critical. To handle this situation, a Best Arm Identification (BAI) Mode (Algorithm 5) is developed based on the new MAB-EA framework.
\begin{algorithm}[tb]
	\caption{Best Arm Identification Mode}
	\label{pseudo_BAIMode}
	\begin{algorithmic}[1]
		\REQUIRE ${}$
		\\ Same control parameters as in Algorithm 4
		\\ $K_e$: Size for the elite pool
		\\ $T_e$: Additional traffic for Best Arm Identification phase
		\STATE $D = P_1$, $D$ is the archive for storing evaluated candidates
		\STATE Initialize elite pool $E$ as empty
		\FOR{$g = 1$ to $G_{\mathrm{max}}$}
		\STATE Same as lines 3-6 in Algorithm 4
		\STATE Add the best $C_e$ percentile candidates of current generation to elite pool $E$
		\WHILE{Size of $E$ is larger than $K_e$}
		\STATE Remove the worst candidate from $E$
		\ENDWHILE
		\STATE Create parent pool $A_g$ as the best $C_p$ percentile candidates in current generation
		\STATE Initialize offspring pool $O_g$ as empty
		\WHILE{Size of $O_g$ is less than $K$}
		\STATE Same as lines 11-17 in Algorithm 4
		\ENDWHILE
		\STATE $P_{g+1} = O_g$
		\ENDFOR
		\STATE Perform pure exploration MAB algorithm on $E$ with a traffic budget of $T_e$, return the identified best candidate
	\end{algorithmic}
\end{algorithm}

In BAI mode, an additional BAI phase is applied after the evolution process has concluded. A MAB algorithm for pure exploration (e.g., SR algorithm) will be performed on an elite archive, i.e., the collection of top candidates over all generations. A single winner will be returned after the BAI phase. Although additional traffic is needed for running the BAI phase, this cost can be compensated by extracting a small portion of traffic from each previous generation (e.g., 10\%). Empirical tests in later section show that BAI mode can significantly improve the reliability of identified best candidate without incurring any additional cost. 

One additional modification in BAI mode is the removal of elite survival mechanism. No candidate is allowed to survive for more than one generation, and all the candidates for next generation will be totally new. The purpose for this modification is to further improve the explorative ability of the framework, considering the fact that the evaluations are very expensive and only limited number of generations (less than 10) are acceptable in real CRO cases. Since the elite archive in BAI mode has already stored the outstanding candidates for every generation, the evolution can focus on exploring more regions in search space.
\subsection{Campaign Mode with Asynchronous Multi-armed Bandit Algorithm}
In some scenaria, the goal for CRO is to make the overall conversion rate during optimization as high as possible instead of returning a single winner. To fill this need, a Campaign mode based on MAB-EA is developed by introducing a new concept to existing MAB algorithms: asynchronous statistics.
\begin{algorithm}[tb]
	\caption{Campaign Mode with Asynchronous MAB Algorithm}
	\label{pseudo_BasicFramework}
	\begin{algorithmic}[1]
		\REQUIRE ${}$
		\\ Same control parameters as in Algorithm 4 excluding $C_e$
		\STATE Initialize the total number of conversions $s_i$ and total number of visits $n_i$ for each candidate as 0
		\FOR{$g = 1$ to $G_{\mathrm{max}}$}
		\STATE Perform asynchronous MAB algorithm on $P_g$ with a traffic budget of $T$, update the total number of conversions $s_i$ and total number of visits $n_i$ for each candidate
		\STATE Same as lines 4-6 in Algorithm 4
		\STATE Create parent pool $A_g$ as the best $C_p$ percentile candidates in current generation
		\STATE Remove the worst $C_p$ percentile candidates from $P_g$
		\STATE Initialize offspring pool $O_g$ as empty
		\WHILE{Size of $P_g + O_g$ is less than $K$}
		\STATE Same as lines 11-13 in Algorithm 4
		\IF{the offspring is not in $P_g + O_g$}
		\STATE Initialize the total number of conversions $s_i$ and total number of visits $n_i$ for the offspring as 0
		\STATE Add the offspring to $O_g$
		\ENDIF
		\ENDWHILE
		\STATE $P_{g+1} = P_g + O_g$
		\ENDFOR
	\end{algorithmic}
\end{algorithm}

The original MAB algorithms initialize the statistics (total reward, average reward, number of pulls, etc.) of all the arms as 0. In contrast, the MAB algorithms in Campaign mode run in an asynchronous manner: All the candidates surviving from previous generation preserve their statistics and use them to initialize the MAB algorithm, which then updates them further, as usual. Taking asynchronous TS as an example, each candidate has an $S_i$ and $F_i$, and these two numbers are updated over generations until the candidate fails to survive. The underlying rationale is that preservation of statistics increases the survival probability of good candidates, therefore Campaign mode focuses more on exploitation than exploration. Asynchronous MAB algorithms allocate more traffic to the existing elites without reevaluating them from scratch, thus improving overall conversion rate. Algorithm 6 summarizes the structure of Campaign mode.
\begin{figure*}[t]
	\centering
	\includegraphics[width=7.0in]{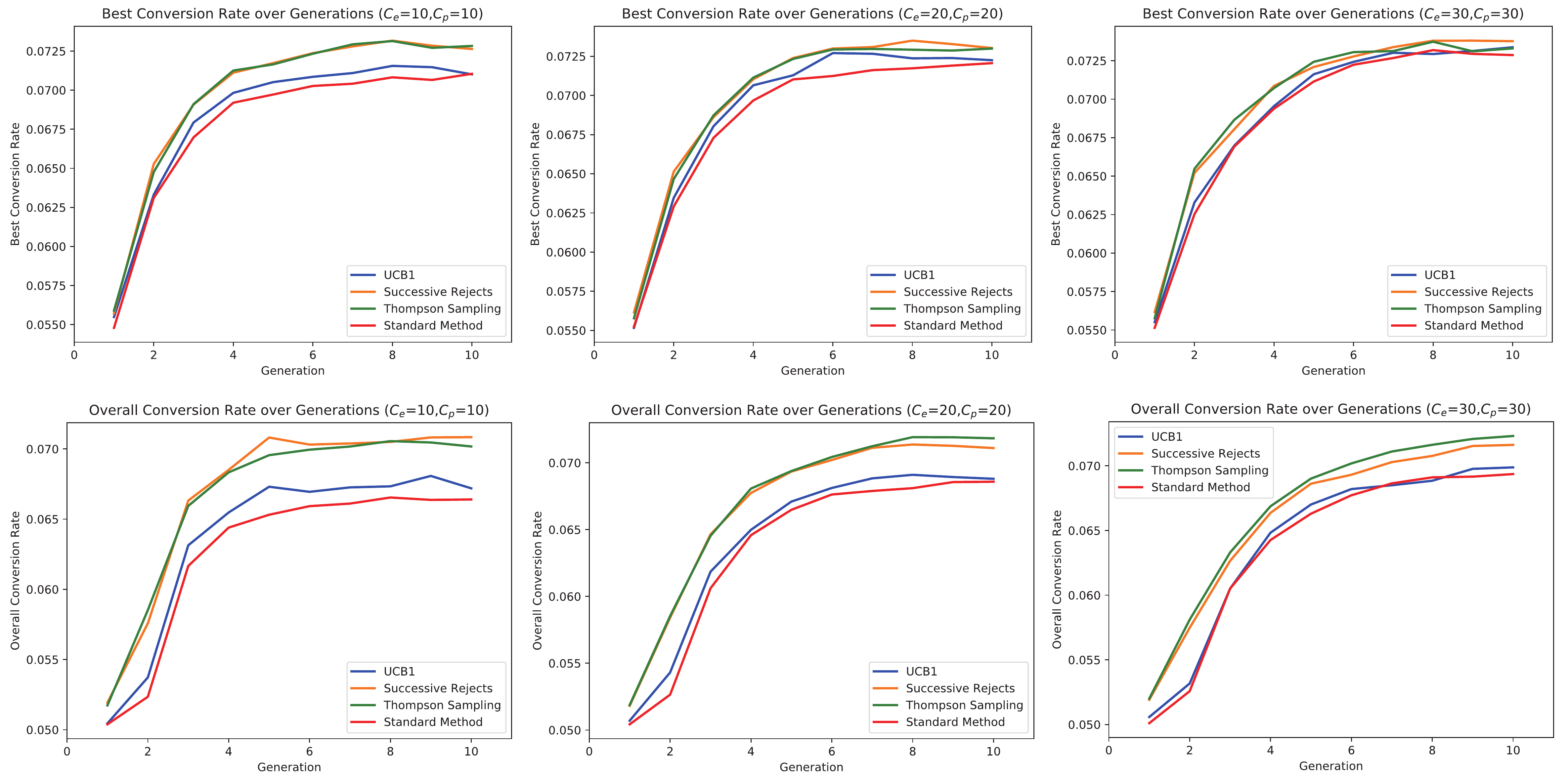}
	\DeclareGraphicsExtensions.
	\vspace*{-7mm}
	\caption{The figure shows the best conversion rate and overall conversion rate in each generation. The results are averaged over 500 independent runs for different $C_e$ and $C_p$ settings. TS and SR perform significantly better than Standard Method in terms of both measures. The differences in best conversion rate are statistically significant after Generation 1, and the differences in overall conversion rate are statistically significant over all generations.}
	\label{BasicFramework}
\end{figure*}

Except for asynchronous statistics in MAB algorithms, the Campaign mode differs from the original MAB-EA in two other aspects. First, the duplication avoidance mechanism is weakened. Since exploration is not the first priority in Campaign mode, duplications between different generations are allowed to encourage the revival of underestimated candidates. Second, only the worst $C_p$ percentile candidates are replaced by the new offspring, which are generated from the top $C_p$ percentile candidates. By setting $C_p$ as 20 or even less, the portion of newly generated offspring is limited, and the overall conversion rate is more stable. Moreover, because all the offspring are generated based on the top candidates, the overall quality of offspring tends to be better than purely random sampling. Under these mechanisms, the crossover and mutation operations can continue the exploration at a steady pace.
\section{Empirical Study}
This section evaluates the proposed framework and mechanisms via experiments based on real-world data. All the conclusions in this section are supported by $t$-test at a 5\% significance level.
\subsection{Experimental Setup}
The Sentient Ascend system contains a simulator that allows testing conversion rate performance on simulated traffic. It therefore makes it possible to evaluate the effect of new algorithms in formal and controlled conditions before they are applied to real traffic. There are multiple possible choices for each element of a website, and each choice will increase or decrease the basic conversion rate of the website. The effect of each choice is predefined and kept fixed during the CRO process. For all the experiments in this section, the following setup is used: There are a total of 8 elements in the website that needs to be optimized. The elements have 5, 4, 2, 3, 4, 3, 3 and 4 choices, respectively. The basic conversion rate for the website is 0.05, and the effect of each element choice is within $[-0.01, 0.01]$. The mean conversion rate for all possible designs is 0.04997, and the conversion rate for the best possible design is 0.08494. This parametric setup is based on real-world data. For each simulated visit, a Bernoulli test with success probability equal to the conversion rate of the web design will be conducted. A successful trial corresponds to a successful conversion, and givies a reward of 1. A failed trial returns a reward of 0.
\subsection{Overall Performance Evaluation}
To evaluate the performance of the new framework, three representative MAB algorithms (SR, TS and UCB1) are incorporated into it, and an empirical comparison between these three variants and the original evolutionary CRO algorithm is conducted. The original algorithm is the same as Algorithm 4 except for traffic allocation: instead of varying it based on a MAB algorithm, all candidates evenly share the traffic budget. For convenience, the original evolutionary CRO algorithm is named "Standard Method" in the rest of the paper. 

The traffic budget for each generation is fixed at 10,000, and the maximun number of generations is set at 10 , conforming to cost limitations in real-world CRO. The population size $K$ is 20, and the mutation probability $C_m$ is 0.01. Different values of elite and parent percentages, $C_e$ and $C_p$, are tested to investigate the robustness of the proposed framework. Two performance metrics are utilized: one is the best conversion rate, i.e., the true conversion rate of the best-performing candidate in each generation; the other is the overall conversion rate for each generation, i.e., the total number of conversions in one generation divided by the total number of visits in that generation. Note that the overall conversion rate is different from simply averaging the conversion rates of all the candidates, because the traffic allocated to each candidate may be different. 

Figure 1 shows the results based on 500 independent runs. From Figure 1, it is clear that the proposed framework with TS and SR significantly increases the overall conversion rate during evolution without deteriorating the optimization performance. In fact, the incorporation of MAB algorithms even improves the optimization performance in terms of best conversion rate. Regarding the influence of $C_e$ and $C_p$, larger values of these two parameters lead to more explorative behaviors at a cost of overall conversion rate at early stage. In real-world cases, an acceptable generation number is usually 5, so a reasonable choice for $C_e$ and $C_p$ would be 20 or even less. Under these circumstances, the TS and SR variants perform best both in terms of overall conversion rate and best conversion rate. There are three explanations: first, the MAB algorithm allocates more traffic to the promising candidates, thereby increasing the overall conversion rate during evaluation; second, since the top candidates receive more traffic from MAB algorithm, the reliability of best performing candidate is enhanced; third, under small $C_e$ and $C_p$, the quality of offspring relies heavily on the top candidates, and more reliable top candidates tend to generate more reliable offspring. The overall quality of candidates is therefore improved, and the overall conversion rate is further increased in this way. Regarding UCB1, since the average reward in the simulated CRO case is very low (e.g., 0.05), the exploration term ($\sqrt{2 \log{(t)} / n_i}$ in line 7 of Algorithm 1) plays a more important role in arm selection. This encourages evenly allocation of the traffic, thereby leading to similar behaviors as in Standard Method.
\subsection{Effectiveness of Best Arm Identification Mode}
\begin{figure}[t]
	\centering
	\includegraphics[width=3.3in]{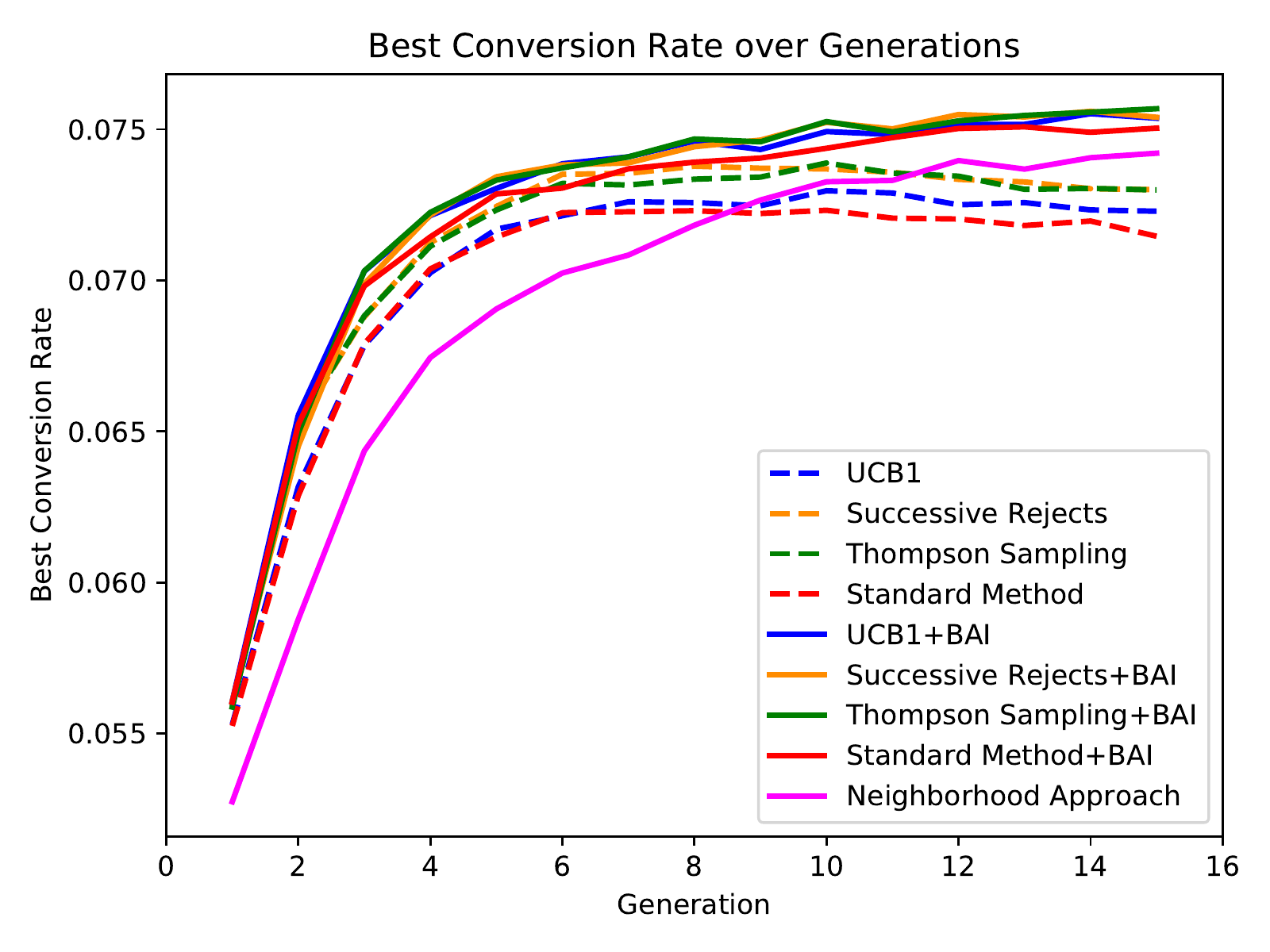}
	\DeclareGraphicsExtensions.
	\vspace*{-8mm}
	\caption{Best conversion rate over generations. The methods with a BAI phase perform significantly better, i.e. they allow identifying a candidate where true performance is significantly better than methods without a BAI phase. The neighborhood approach is better than non-BAI methods in later stages, but not as good as BAI variants. The results are averaged over 500 independent runs, and the performance differences between BAI variants and non-BAI variants are statistically significant.}
	\label{BasicFramework}
\end{figure}
This subsection demonstrates the effectiveness of BAI mode through an experimental comparison with the Standard Method, MAB-EA, and a state-of-the-art approach \cite{miikkulainen:ssci17} where the average fitness within a predefined neighborhood (in solution space) is used to evaluate candidates. In the experiments, the neighborhood approach was further improved by considering all the previous candidates when calculating neighbood fitnesses. 

For a fair comparison, MAB-EA and neighborhood approach have 11,000 visits per generation; BAI mode has 10,000 visits for each generation and 10,000 additional visits in the BAI phase. The other parameters are identical for all algorithms: $C_e = 20$, $C_p = 20$, $K = 20$, $G_{\mathrm{max}} = 15$, $C_m = 0.01$. For BAI mode, $K_e = 20$, and SR algorithm is used in BAI phase. For neighborhood approach, neighborhood size is fixed at 5. 

Figure 2 compares the best conversion rates of all the algorithms averaging over 500 independent runs. BAI mode consistently improves over the Standard Method, MAB-EA, and neighborhood approach. It both converges faster early on, and explores more efficiently later. After Generation 10, BAI mode significantly outperforms MAB-EA even with less total traffic. The neighborhood approach's performance gradually improves with the collection of more candidates. However, BAI mode is still more reliable than the neighborhood approach even in later stages. Based on the experimental results, BAI mode allows selecting a better winner, and estimates its future/true performance more accurately. It therefore provides important improvements in practical applications.
\subsection{Effectiveness of Asynchronous MAB Algorithm in Campaign Mode}
The main difference between Campaign mode and MAB-EA is the new asynchronous MAB algorithm. This section verifies the effectiveness of asynchronous statistics in MAB algorithms via an empirical comparison. In the experiments, SR, TS and UCB1 are modified to run asynchronously and compared with their original versions, as well as with the Standard Method. The same parameters are used for all algorithms: $C_p = 20$, $K = 20$, $C_m = 0.01$, $T = 10,000$. Since Campaign mode usually run for longer, $G_{\mathrm{max}}$ is set at 50. 
\begin{figure}[t]
	\centering
	\includegraphics[width=3.3in]{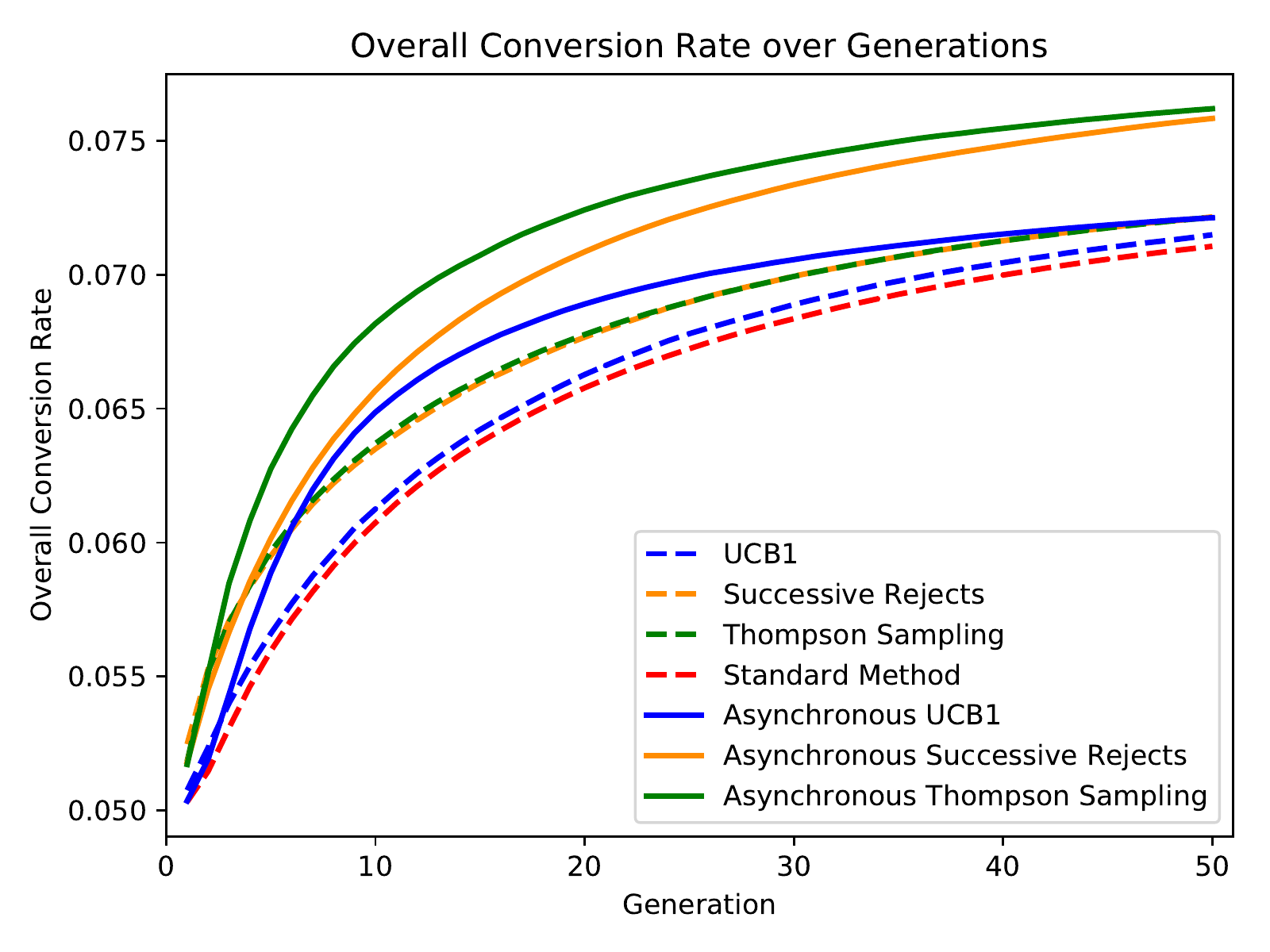}
	\DeclareGraphicsExtensions.
	\vspace*{-8mm}
	\caption{The overall conversion rate for entire optimization process in Campaign Mode. The data point at generation $g$ shows the overall conversion rate until generation $g$. The asynchronous versions of TS and SR perform significantly better than other variants, leading to better conversion rate over the entire campaign. The results are averaged over 500 independent runs, and the performance differences between asynchronous versions and original versions are statistically significant for all tested MAB algorithms.}
	\label{BasicFramework}
\end{figure}

Figure 3 compares the results over 500 independent runs. Asynchronous SR and asynchronous TS perform significantly better than their original versions. For UCB1, the asynchronous version is better only in the early stages. This is because all the candidates in UCB1 algorithm share the parameter $t$ (total number of visits for all the candidates). A candidate that has survived for a long period in the asynchronous variant will lead to a very large $t$. A significant bias towards less visited candidates will then be introduced during the traffic allocation (line 7 of Algorithm 1), thereby wasting more traffic in those unreliable candidates. The candidates in SR and TS do not share any parameters, so the Campaign mode works properly with their asynchronous versions, improving overall conversion rate during optimization significantly.
\section{Discussion and Future Work}
The proposed mechanisms in this work solve three general issues in evolutionary CRO: how to allocate the evaluation budget efficiently, how to make the optimization outcome reliable, and how to maintain high overall conversion rate during evolution. Although the new approaches are only demonstrated in CRO domain, all of them can be generalized to other optimization problems with noisy fitness functions.

The main idea of MAB-EA is to utilize MAB algorithms to allocate the evaluation budget. Although poorly performing candidates cannot receive sufficient traffic to estimate their true fitnesses accurately, the optimization performance does not deteriorate. This is because the evolution pressure in EAs comes from the parent selection and survival selection, and these two steps rely primarily on those good candidates. That is, efficient detection of good candidates is more important than accurate evaluation of bad candidates. Thus, MAB algorithms reduce evaluation cost without sacrificing optimization performance.

The BAI mode is a significant improvement in settings where result reliability is critical. The EA maintains an elite archive that collects good candidates during optimization. After evolution is finished, a pure exploration MAB algorithm is performed on the elite archive to select the final winner. This process amounts to a two-level winner selection, and reliability of the optimization outcome is enhanced. The additional traffic for the BAI phase is extracted from previous generations, thus no extra cost is incurred.

For situations in which overall conversion rate during optimization matters, the new concepts in Campaign mode can be applied. Asynchronous MAB algorithms together with a high survival probability and greedy offspring generation lead to a high yet stable overall conversion rate during evolution. It is notable that only MAB algorithms without sharing parameters among candidates are suitable for asynchronization, such as TS and SR but not UCB1.

One interesting future direction is to introduce contextual bandit algorithms \cite{AgarwalHKLLS14}, in which the interrelations among variables are explicitly modeled. The model will then be used by the MAB algorithms to allocate evaluation budget more efficiently. Moreover, the model can be used in the crossover or mutation operations to propagate promising variable combinations more often, thus increasing overall performance and efficiency further. Another direction is the incorporation of asynchronous statistics into BAI mode. Initializing elites in BAI phase with their statistics in the main optimization phase may further increase the reliability of best candidate.

Given that the simulation results in this paper are so positive, the techniques are currently being implemented in the Sentient Ascend product. Thus, they will be in use in optimization of real-world web interfaces shortly. Beyond CRO, the techniques should be useful in many other domains where fitness evaluations are noisy, such as game playing and robotics, where fitness depends on stochastic interactions with the environment, and neural architecture search, where it depends on stochastic initialization and training. Thus, the techniques can potentially have a high impact on optimization of complex systems.
\section{Conclusion}
This paper demonstrates how MAB algorithms can be used to make EAs more effective in uncertain domains. First, the proposed MAB-EA framework makes it possible to allocate the available evaluation budget more efficiently. Second, the BAI mode, based on a pure exploration MAB algorithms, makes the winner selection more reliable. Third, the Campaign mode, based on asynchronous MAB algorithms, achieves a high and stable overall conversion rate during the entire optimization process.  These mechanisms are shown to be effective in the CRO domain, but should readily extend to other applications of machine discovery in noisy environments, e.g., game playing, robotics, and neural architecture search.

{\footnotesize 
%\section{Acknowledgments}
\bibliographystyle{aaai}
\bibliography{ref}}

\begin{thebibliography}{}

\bibitem[\protect\citeauthoryear{Agarwal \bgroup et al\mbox.\egroup
  }{2014}]{AgarwalHKLLS14}
Agarwal, A.; Hsu, D.~J.; Kale, S.; Langford, J.; Li, L.; and Schapire, R.~E.
\newblock 2014.
\newblock Taming the monster: {A} fast and simple algorithm for contextual
  bandits.
\newblock {\em CoRR} abs/1402.0555.

\bibitem[\protect\citeauthoryear{Agrawal and Goyal}{2012}]{AgrawalG12}
Agrawal, S., and Goyal, N.
\newblock 2012.
\newblock Analysis of thompson sampling for the multi-armed bandit problem.
\newblock In {\em {COLT} 2012, June 25-27, 2012, Edinburgh, Scotland},
  39.1--39.26.

\bibitem[\protect\citeauthoryear{Audibert and Bubeck}{2010}]{Audibert2010}
Audibert, J.-Y., and Bubeck, S.
\newblock 2010.
\newblock Best arm identification in multi-armed bandits.

\bibitem[\protect\citeauthoryear{Auer, Cesa-Bianchi, and
  Fischer}{2002}]{Auer2002}
Auer, P.; Cesa-Bianchi, N.; and Fischer, P.
\newblock 2002.
\newblock Finite-time analysis of the multiarmed bandit problem.
\newblock {\em Machine Learning} 47(2):235--256.

\bibitem[\protect\citeauthoryear{Bubeck, Munos, and Stoltz}{2009}]{Bubeck2009}
Bubeck, S.; Munos, R.; and Stoltz, G.
\newblock 2009.
\newblock {\em Pure Exploration in Multi-armed Bandits Problems}.
\newblock Berlin, Heidelberg: Springer Berlin Heidelberg.
\newblock  23--37.

\bibitem[\protect\citeauthoryear{Chapelle and Li}{2011}]{Chapelle2011}
Chapelle, O., and Li, L.
\newblock 2011.
\newblock An empirical evaluation of thompson sampling.
\newblock NIPS'11,  2249--2257.
\newblock USA: Curran Associates Inc.

\bibitem[\protect\citeauthoryear{Chernoff}{1952}]{Chernoff1952}
Chernoff, H.
\newblock 1952.
\newblock A measure of asymptotic efficiency for tests of a hypothesis based on
  the sum of observations.
\newblock {\em The Annals of Mathematical Statistics} 23(4):493--507.

\bibitem[\protect\citeauthoryear{Eiben and Smith}{2015}]{Nature2015}
Eiben, A.~E., and Smith, J.
\newblock 2015.
\newblock From evolutionary computation to the evolution of things.
\newblock {\em Nature} 521:476 EP --.

\bibitem[\protect\citeauthoryear{Granmo}{2010}]{Ole2010}
Granmo, O.
\newblock 2010.
\newblock Solving two‐armed bernoulli bandit problems using a bayesian
  learning automaton.
\newblock {\em International Journal of Intelligent Computing and Cybernetics}
  3(2):207--234.

\bibitem[\protect\citeauthoryear{Hoeffding}{1963}]{Hoeffding1963}
Hoeffding, W.
\newblock 1963.
\newblock Probability inequalities for sums of bounded random variables.
\newblock {\em Journal of the American Statistical Association} 58(301):13--30.

\bibitem[\protect\citeauthoryear{Kamiura and Sano}{2017}]{KAMIURA201725}
Kamiura, M., and Sano, K.
\newblock 2017.
\newblock Optimism in the face of uncertainty supported by a
  statistically-designed multi-armed bandit algorithm.
\newblock {\em Biosystems} 160(Supplement C):25 -- 32.

\bibitem[\protect\citeauthoryear{Kaufmann, Korda, and
  Munos}{2012}]{Kaufmann2012}
Kaufmann, E.; Korda, N.; and Munos, R.
\newblock 2012.
\newblock Thompson sampling: An asymptotically optimal finite-time analysis.
\newblock ALT'12,  199--213.
\newblock Berlin, Heidelberg: Springer-Verlag.

\bibitem[\protect\citeauthoryear{Miikkulainen \bgroup et al\mbox.\egroup
  }{2017a}]{Miikkulainen2017}
Miikkulainen, R.; Iscoe, N.; Shagrin, A.; Cordell, R.; Nazari, S.; Schoolland,
  C.; Brundage, M.; Epstein, J.; Dean, R.; and Lamba, G.
\newblock 2017a.
\newblock Conversion rate optimization through evolutionary computation.
\newblock GECCO '17,  1193--1199.
\newblock New York, NY, USA: ACM.

\bibitem[\protect\citeauthoryear{Miikkulainen \bgroup et al\mbox.\egroup
  }{2017b}]{miikkulainen:ssci17}
Miikkulainen, R.; Shahrzad, H.; Duffy, N.; and Long, P.
\newblock 2017b.
\newblock How to select a winner in evolutionary optimization?
\newblock SSCI '17.
\newblock IEEE.

\bibitem[\protect\citeauthoryear{Miikkulainen \bgroup et al\mbox.\egroup
  }{2018}]{miikkulainen:iaai18}
Miikkulainen, R.; Iscoe, N.; Shagrin, A.; Rapp, R.; Nazari, S.; McGrath, P.;
  Schoolland, C.; Achkar, E.; Brundage, M.; Miller, J.; Epstein, J.; and Lamba,
  G.
\newblock 2018.
\newblock Sentient ascend: {AI}-based massively multivariate conversion rate
  optimization.
\newblock IAAI '18.
\newblock AAAI.

\bibitem[\protect\citeauthoryear{Robbins}{1952}]{robbins1952}
Robbins, H.
\newblock 1952.
\newblock Some aspects of the sequential design of experiments.
\newblock {\em Bull. Amer. Math. Soc.} 58(5):527--535.

\bibitem[\protect\citeauthoryear{Salehd and Shukairy}{2011}]{Khalid11}
Salehd, K., and Shukairy, A.
\newblock 2011.
\newblock {\em Conversion Optimization: The Art and Science of Converting
  Prospects to Customers}.
\newblock Sebastopol, CA: O'Reilly Media, Inc.

\bibitem[\protect\citeauthoryear{Scott}{2010}]{Scott2010}
Scott, S.~L.
\newblock 2010.
\newblock A modern bayesian look at the multi-armed bandit.
\newblock {\em Appl. Stoch. Model. Bus. Ind.} 26(6):639--658.

\bibitem[\protect\citeauthoryear{Thompson}{1933}]{Thompson1933}
Thompson, W.~R.
\newblock 1933.
\newblock On the likelihood that one unknown probability exceeds another in
  view of the evidence of two samples.
\newblock {\em Biometrika} 25(3/4):285--294.

\bibitem[\protect\citeauthoryear{Weber}{1992}]{Richard92}
Weber, R.
\newblock 1992.
\newblock On the gittins index for multiarmed bandits.
\newblock {\em The Annals of Applied Probability} 2(4):1024--1033.

\end{thebibliography}

\end{document}